\documentclass[11pt,a4paper]{article}
\usepackage[hyperref]{emnlp-ijcnlp-2019}
\usepackage{times}
\usepackage{latexsym}
\usepackage{array}
\usepackage{amsmath}
\usepackage{url}
\usepackage{verbatim}
\usepackage{graphicx}
\usepackage{microtype}

\aclfinalcopy 

\title{Evaluating Language Model Finetuning Techniques for Low-resource Languages}

\author {Jan Christian Blaise Cruz \textnormal{and} Charibeth Cheng \\
  Center for Language Technologies \\
  College of Computer Studies \\
  De La Salle University, Manila \\
  {\tt \{jan\_christian\_cruz, charibeth.cheng\}@dlsu.edu.ph} \\}

\date{}

\begin{document}
\maketitle
\begin{abstract}
  Unlike mainstream languages (such as English and French), low-resource languages often suffer from a lack of expert-annotated corpora and benchmark resources that make it hard to apply state-of-the-art techniques directly. In this paper, we alleviate this scarcity problem for the low-resourced Filipino language in two ways. First, we introduce a new benchmark language modeling dataset in Filipino which we call WikiText-TL-39. Second, we show that language model finetuning techniques such as BERT and ULMFiT can be used to consistently train robust classifiers in low-resource settings, experiencing at most a 0.0782 increase in validation error when the number of training examples is decreased from 10K to 1K while finetuning using a privately-held sentiment dataset. 
\end{abstract}

\section{Introduction}
The use of neural networks in Natural Language Processing (NLP) has achieved great successes in multiple areas such as language modeling \cite{merity2016Pointer}, machine translation \cite{vaswani2017attention, bahdanau2014neural}, and multitask learning \cite{radford2019language, mccann2018natural}. 

While effective, neural network methods are data-hungry and do not operate well in data scarce settings such as with low-resource languages \cite{zoph2016transfer}. In addition, such languages may also not have readily-available resources found in mainstream languages such as pretrained word embeddings and expert-annotated corpora \cite{adams2017cross}. 

This data scarcity problem is best met with the construction of properly annotated corpora for such tasks, however such annotation work is cost-prohibitive and time-consuming \cite{cotterell2017low}. Techniques must be developed to address the low-resource case in NLP and allow robust models to be trained despite data scarcity \cite{cotterell2017low}.

Transfer learning provides one way to offset this data scarcity problem, allowing models to be pretrained then suibsequently finetuned on a smaller dataset, reducing not only the resource requiremens, but also the compute and time requirements to achieve a robust model \cite{howard2018universal}.

In this paper, we provide two contributions: first, we release the first open, large-scale preprocessed unlabeled text corpora in the low-resource Filipino language which we call ``WikiText-TL-39.'' Second, we show that transfer learning techniques such as BERT \cite{devlin2018bert} and ULMFiT \cite{howard2018universal} can be used to train robust classifiers in low-resource settings, experiencing at most a 0.0782 increase in error when the number of training examples is reduced from 10K to 1K. 

We open source all pretrained models and datasets in an open, public repository\footnote{https://github.com/jcblaisecruz02/Tagalog-BERT}. 

\section{Methodology}
Our evaluation methodology is as follows: First, we construct a large-scale unlabeled text corpora to train pretrained language models to transfer from. Second, we evaluate transfer learning performance on a privately held sentiment dataset. We will then steadily decrease the number of training examples and study the changes on validation accuracy.

We use two transfer learning techniques, namely BERT \cite{devlin2018bert} and \cite{howard2018universal}.

\begin{figure*}[ht]
    \includegraphics[width=1\textwidth]{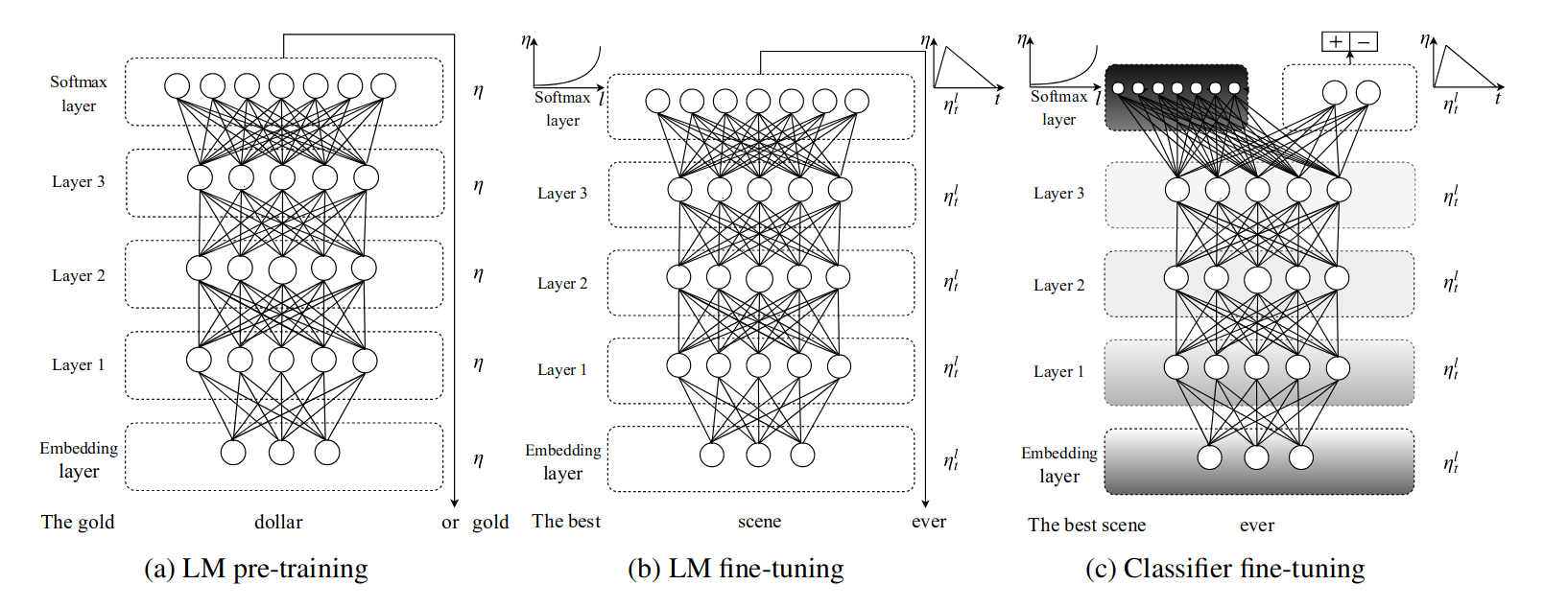}
    \caption{Overall ULMFiT pretraining and finetuning framework. An AWD-LSTM \cite{awdlstm2017merity} is pretrained on a language modeling task. The weights are then reused with no modifications to the architecture. For finetuning, the model is first finetuned, again using language modeling, this time to the text of the target dataset to adapt to its own vocabulary and idiosyncracies. Lastly, a ``classification layer'' is added to the model and is finetuned for text classification. Adapted from \citet{howard2018universal}.}
    \label{fig:ulmfit}
\end{figure*}

\subsection{ULMFiT}

ULMFiT \cite{howard2018universal} was introduced as a transfer learning method for Natural Language Processing that works akin to ImageNet \cite{imagenet2015russakovsky} pretraining in Computer Vision. 

It uses an AWD-LSTM \cite{awdlstm2017merity} pretrained on a language modeling objective as a base model, which is then finetuned to a downstream task in two steps. 

First, the language model is finetuned to the text of the target task to adapt to it syntactically. Second, a classification layer is appended to the model and is finetuned to the classification task conservatively. During finetuning, multiple different techniques are introduced to prevent catastrophic forgetting, wherein the model loses most (if not all) information and relations it has learned during the pretraining stage. 

ULMFiT holds state-of-the-art for text classification, and is notable for being able to set comparable scores with as little as 1000 samples of data which makes it attractive for use in low-resource settings.

An overview schematic of ULMFiT can be found in figure \ref{fig:ulmfit}.

\begin{figure*}[ht]
    \includegraphics[width=1\textwidth]{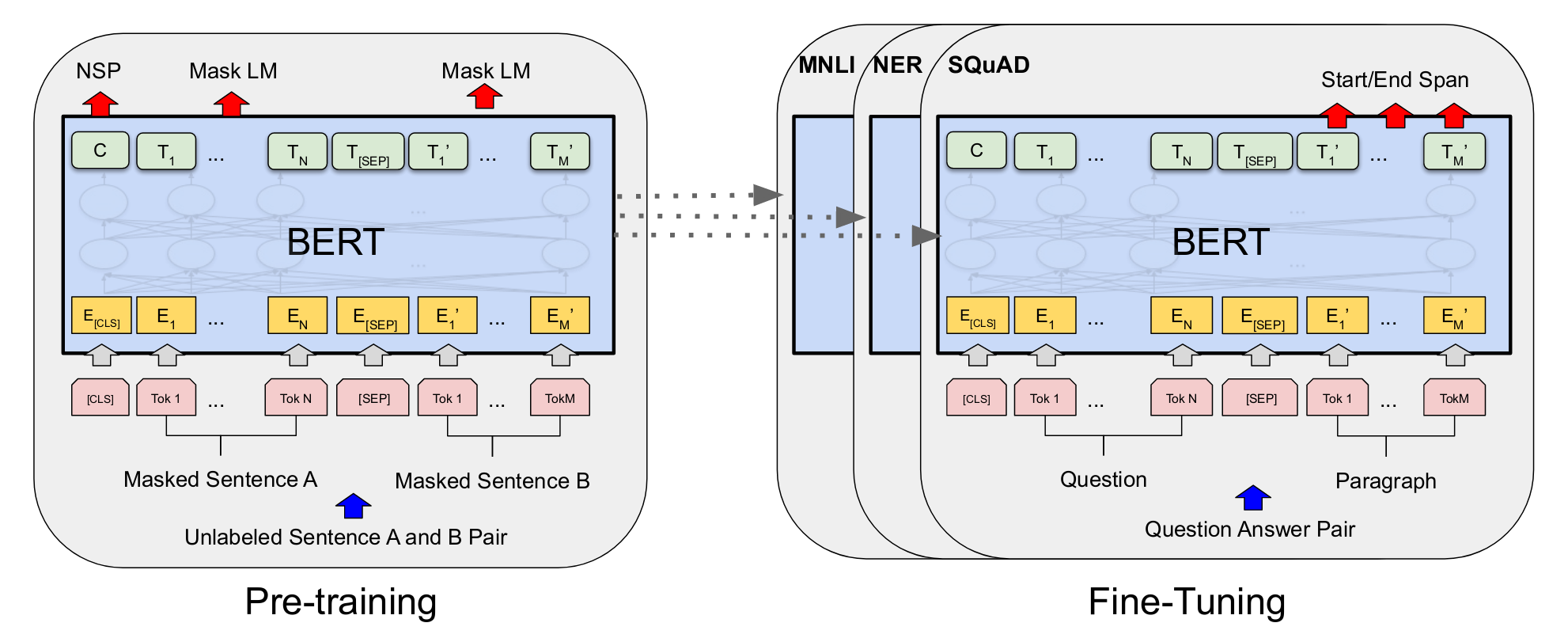}
    \caption{Overall BERT pretraining and finetuning framework. Note that the same architecture in pretraining is also used in finetuning with little-to-no modification in structure. After masked-language model and next-sentence prediction is pretrained, we transfer the weights of the model to downstream tasks, with question answering and entailment shown in this example. Adapted from \citet{devlin2018bert}.}
    \label{fig:bert}
\end{figure*}

\subsection{BERT}

BERT is a transformer-based \cite{vaswani2017attention} language model that is designed to pretrain ``deep bidirectional representations'' that can be finetuned to different tasks, with state-of-the-art results achieved in multiple benchmarks \cite{devlin2018bert}. 

BERT's power comes from Attention, a mechanism that allows a network to give more weight to certain tokens in a sequence, essentially ``paying more attention to important parts'' \cite{vaswani2017attention}. Precisely, we compute attention on a set of queries packed as a matrix $Q$ on key and value matrices $K$ and $V$, respectively, as:
\begin{equation}
    \begin{aligned}
        & \textnormal{Attention}(Q, K, V) = \textnormal{softmax}(\frac{QK^{T}}{\sqrt{d_{k}}})V
    \end{aligned}
\end{equation}
where $d_{k}$ is the dimensions of the key matrix $K$. Attention allows BERT to model not only sequences, but also the importance and weight of each token in a sequence with respect to other sequences, as well as itself. 

In addition to leveraging Attention, it uses the Transformer \cite{vaswani2017attention} architecture, to where BERT gains its bidirectionality. Transformers are sequence models that do not use recurrent layers, instead leveraging only feed-forward layers and attention mechanisms. The disuse of recurrences provide two advantages: First, it allows transformers to be parallelized as they are not sequential in nature unlike LSTMs or GRUs. Second, they allow batches of text to be seen at once, again due to its unsequential nature, which also in turn allows it to leverage attention mechanisms and be bidirectional.

 BERT is unique that it uses modified tasks for pretraining. Given that its bidirectionality gives it access to left-context, the model would be able to ``peek'' directly at the next words when following a standard language modeling task. To alleviate this, the authors propose the use of ``masked-language modeling,'' which masks a number of words in the sentence with the model tasked to identify them \cite{devlin2018bert}. In addition, a second pretraining task called ``next-sentence prediction'' was added to enforce stronger relationships between two sentences. In this task, a target sentence is identified if it is likely to precede a source sentence \cite{devlin2018bert}. 

In addition to these augmentations, BERT also benefits from being \textit{deep}, allowing it to capture more context and information. BERT-Base, the smallest BERT model, has 12 layers (768 units in each hidden layer) and 12 attention heads for a total of 110M parameters. Its larger sibling, BERT-Large, has 24 layers (1024 units in each hidden layer) and 16 attention heads for a total of 340M parameters.

An overview schematic of BERT can be found in figure \ref{fig:bert}.

\section{WikiText-TL}
A difficulty in adapting pretraining methods to low-resource languages is the lack of processed datasets large enough to train robust pretrained models. Inspired by the original WikiText Long Term Dependency Language Modeling Dataset \cite{merity2016Pointer}, we introduce a benchmark dataset which we call WikiText-TL-39, where ``TL'' stands for \textit{Tagalog} and ``39'' refers to the dataset having 39 million tokens in the training set. The corpus statistics for WikiText-TL-39 is shown on table \ref{corpusstats}.

\begin{table*}[t!]
\centering
\begin{tabular}{lllll}
\hline
  Split & Documents & Tokens & Unique Tokens & Num. of Lines \\ \hline
  Training & 120,975 & 39,267,089 & 279,153 & 1,403,147 \\
  Validation & 25,919 & 8,356,898 & 164,159 & 304,006 \\
  Testing & 25,921 & 8,333,288 & 175,999 & 298,974 \\
  \hline
  OOV Tokens & 28,469 (0.1020\%) & & \\
  \hline
  
\end{tabular}
\caption{\label{corpusstats} Statistics for the WikiText-TL-39 Dataset.}
\end{table*}

\subsection{Construction and Pre-processing}
Since Tagalog Wikipedia does not have a list of verified ``good'' articles \cite{merity2016Pointer} and has far fewer content pages unlike its English counterpart (5,800,000 in English vs. 75,000), we opted to instead scrape the content from all the listed pages in the Tagalog Wikipedia table of contents\footnote{https://tl.wikipedia.org/wiki/Natatangi:Lahat\_ng\_mga\_pahina}, narrowing down to just articles with titles that start with letters A-Z. Content was extracted using open-source Python packages Requests\footnote{https://pypi.org/project/requests/} and BeautifulSoup\footnote{https://pypi.org/project/beautifulsoup4/}.

All characters were normalized into unicode and all HTML markup were unescaped. Normalization and tokenization were performed via the Moses Tokenizer \cite{koehn2007Moses}. We split the corpus into training, validation, and test sets with a ratio of 70\%-15\%-15\%, respectively. When constructing the vocabulary, we opted to not discard words that had a vocabulary count of less than 3, unlike in \cite{chelba20131B}. This resulted in a vocabulary size of 279,153 tokens. We replace all tokens in the test set unseen in the training set with special \textless unk\textgreater tokens.

\subsection{Model-specific Pre-processing}
Pretraining with BERT requires a trained WordPiece vocabulary. We opted to use the Byte-Pair Encoding (BPE) \cite{sennrich2016BPE} model in Google's SentencePiece\footnote{https://github.com/google/sentencepiece} library to train our own vocabulary as Google did not release the original WordPiece code due to it having dependencies with their own internal libraries.

We experiment with two fixed vocabulary sizes in pretraining BERT. We generate a vocabulary with 290,000 tokens, following the original vocabulary size of the dataset. We also generate a vocabulary with a fixed size of 30,000 tokens, following the original specifications of Google's own pretrained English BERT models\footnote{https://github.com/google-research/bert}.

For use in ULMFiT, we followed a light preprocessing scheme that involves converting all words to lowercase, with a special \textless maj \textgreater token added in front of words that originally start with a capital letter. We likewise change all unknown words to the \textless unk \textgreater token, and limit the vocabulary to the top 30K words.

\section{Experiments}
\subsection{BERT Pretraining}
We pretrain BERT Base models with 12 layers, 768 neurons per hidden layer, and 12 attention heads (a total of about 110M parameters) on our prepared corpus and SentencePiece vocaularies using Google's provided pretraining scripts\footnote{https://github.com/google-research/bert}.

We experiment by varying the casing (cased and uncased models), the vocabulary size (full 290K vs 30K), and the number of training and warmup steps (1M steps with 10K warmups and 500K steps with 5K warmups). 

For the masked language model pretraining objective, we follow the original specifications and use a 0.15 probability of a word being masked. We also set the maximum number of masked language model predictions to the original 20. All models use a maximum sequence length of 128 and a batch size of 256. We use a learning rate of 1e-4 for all models.

All models are pretrained on Google Cloud Compute Engine using Google's Tensor Processing Units (TPU) version 2.8.

\subsection{AWD-LSTM Pretraining}
For ULMFiT, we train an AWD-LSTM language model using our prepared corpus. We train a 3-layer model and use an embedding size of 400 and a hidden size of 1150. We set the dropout values for the embedding, the RNN input, the hidden-to-hidden transition, and the RNN output to (0.1, 0.3, 0.3, 0.4) respectively. We use a weight dropout of 0.5 on the LSTM's recurrent weight matrices.

The model was trained for 30 epochs with a learning rate of 1e-3, a batch size of 128, and a weight decay of 0.1. We use the Adam optimizer and use  slanted triangular learning rate schedules. We train the model on a machine with one NVIDIA Tesla V100 GPU. 

\subsection{Sentiment Classification Task}
We finetune on a privately held sentiment classification dataset containing 10K positive and 10K negative reviews on electronic products.

To simulate low-resource settings, we randomly sample splits from the original dataset: a full 10K-10K split of positive and negative reviews, a 5K-5K split, a 1K-1K split, and a 100-100 split. For both BERT and ULMFiT, we finetune the pretrained models to each split to evaluate performance given the scarcity of data.

To evaluate the performance, we use a validation set of 1500 positive and 1500 negative reviews from the same source. For each split, we use the same validation split without reducing it. This ensures consistency when evaluating the changes in validation accuracy once the number of training examples is reduced.

The dataset is lightly preprocessed using the Moses tokenizer \cite{koehn2007Moses}, keeping casing and placing spaces around punctuation. Contractions with an apostrophe (ie. cannot $\rightarrow$ can't) are not given special tokens nor are preprocessed further as such contractions are rare in Filipino.

\subsection{Finetuning}
For BERT, we finetune our best cased and uncased BERT models on each sentiment classification split. For each finetuning setup, we finetune for 3 epochs with a learning rate of 2e-5. We use a maximum sequence length of 128 and a batch size of 32. 

For ULMFiT, we finetune our pretrained AWD-LSTM language on each of the sentiment classification splits. We first perform language model finetuning with the sentiment classification dataset for 10 epochs, using a learning rate 1e-2. For the original 10k-10k split, we use weight decay of 0.1 and a batch size of 80, and for all other splits we use weight decay of 0.3 and a batch size of 40. We use this final language model to finetune a sentiment classification model in the final stage of ULMFiT.

For the final ULMFiT stage, we finetune via gradual unfreezing. We finetune for five epochs, gradually unfreezing the last layer until all layers are unfrozen on the fourth epoch. We use a learning rate of 1e-3 and set Adam's $\alpha$ and $\beta$ parameters to 0.8 and 0.7 respectively.

We then evaluate on a fixed validation set and record changes in the model performance.

\section{Results and Discussion}
\subsection{Pretraining Results}
For BERT pretraining, we were able to train eight models, varying across vocabulary size, casing, and pretraining steps. Our best uncased model (reaching a final loss of 0.0935) was trained for 500K steps with 5K steps of finetuning on the smaller 30K SentencePiece vocabulary. The best cased model (reaching a final loss of 0.0642), on the other hand, needed 1M pretraining steps with 10K warmup steps on the same 30K SentencePiece vocabulary. We surmise that this is due to the model needing more steps to learn and get accustomed to casing. 

The full results of BERT pretraining can be found on Table \ref{bertpretraining}.

For ULMFiT, our AWD-LSTM language model reached a final validation loss of 4.4857 (which equals to 1.5009 perplexity). The model finished training for 30 epochs after around 11 hours.

\begin{table*}[t!]
\centering
\begin{tabular}{lllllll}
\hline
  Steps / Warmup & Casing & Vocab Size & Loss & MLM Acc & NSP Acc & Train Time \\ \hline
  500K / 5K & Cased & 290K & 0.3198 & 0.9158 & 0.9950 & 22H \\
  500K / 5K & Cased & 30K & 0.1046 & 0.9865 & 1.0000 & 33H \\
  500K / 5K & Uncased & 290K & 0.3396 & 0.9176 & 0.9986 & 24H \\
  \textbf{500K / 5K} & \textbf{Uncased} & \textbf{30K} & \textbf{0.0935} & \textbf{0.9862} & \textbf{1.0000} & \textbf{33H} \\
  \hline
  1M / 10K & Cased & 290K & 0.1607 & 0.9563 & 0.9988 & 44H \\
  \textbf{1M / 10K} & \textbf{Cased} & \textbf{30K} & \textbf{0.0642} & \textbf{0.9971} & \textbf{1.0000} & \textbf{66H} \\
  1M / 10K & Uncased & 290K & 0.0716 & 0.9965 & 1.0000 & 168H \\
  1M / 10K & Uncased & 30K & 0.2600 & 0.9426 & 1.0000 & 22H \\
  \hline
  
\end{tabular}
\caption{\label{bertpretraining} BERT Pretraining Results. MLM Acc refers to Masked Language Modeling objective accuracy. NSP Acc refers to Next Sentence Prediction objective accuracy. Figures in \textbf{bold} pertain to the best performing cased and uncased models.}
\end{table*}

\subsection{Finetuning Results}

For BERT finetuning, the uncased model performed marginally better than the cased model with a 0.006 increase in accuracy when finetuning on the original 10K-10K split. We can see that when we reduce the training examples from 10K to 1K, we incur at most a 0.0617 increase of error in the cased models, and a 0.0954 increase of error in the uncased models. The error significantly increases once the number of training examples drop to the 100-100 split, with an increase of 0.1484 error in the cased model, and an increase of 0.2554 error in the uncased model. 

When evaluating on the validation set of the original 10K-10K split, we can see similar results as with evaluating on the validation set of each respective split. For the cased models, we only incur a 0.038 increase of error when finetuning on the 1K-1K split, and a 0.23 increase of error when finetuning on the 100-100 split. For the uncased models, we get a 0.0437 and 0.248 increase of error on the 1K-1K split and 100-100 split, respectively. 

The full results of BERT finetuning can be found on table \ref{finetuning}.

For ULMFiT finetuning, our best model was unsurprisingly the one finetuned on the entire 10K-10K split, getting a final validation accuracy of 0.9018. Reducing the number of examples down to the 1K-1K split incurred only a 0.0835 increase in error. On the 100-100 split, on the other hand, we can see that the error increased by a very large margin of 0.4628, reducing the accuracy from 0.9018 to 0.4390. 

Like in the BERT finetuning setups, we can see that the finetuned classifiers give consistently robust results even when evaluated on the larger 10K-10K split validation set. We can see that reducing the examples down to the 1K-1K split increases error by 0.0782, comparable to evaluating on the 1K-1K split's validation set. Likewise, we suffer a large increase in error of 0.4114 when evaluating on the 100-100 split. 

The full results of ULMFiT finetuning can be found on table \ref{finetuning}.

\begin{table*}[t!]
\centering
\begin{tabular}{cccccccc} 
 \hline
 Model Type & Splits & Val Loss & Val Acc & 10K Val Acc & Err Increase & 10K Err Increase \\ \hline
  BERT-Cased & 10k-10k & 0.3492 & 0.8817 & - & - & - \\
  BERT-Cased & 5k-5k & 0.3841 & 0.8760 & 0.8976 &  +0.0057 & -0.0159*\\
  BERT-Cased & 1k-1k & 0.4746 & 0.8200 & 0.8437 & +0.0617 & +0.0380 \\
  BERT-Cased & 100-100 & 0.6122 & 0.7333 & 0.6517 & +0.1484 & +0.2300 \\
  \hline
  BERT-Uncased & 10k-10k & 0.3401 & 0.8887 & - & - & - \\
  BERT-Uncased & 5k-5k & 0.3727 & 0.8793 & 0.8970 & +0.0094 & -0.0083* \\
  BERT-Uncased & 1k-1k & 0.5667 & 0.7933 & 0.8450 & +0.0954 & +0.0437 \\ 
  BERT-Uncased & 100-100 & 0.6606 & 0.6333 & 0.6407 & +0.2554 & +0.2480 \\
 \hline
  ULMFiT & 10k-10k & 0.2496 & 0.9018 & - & - & - \\
  ULMFiT & 5k-5k   & 0.2489 & 0.8961 & 0.8887 & +0.0057 & +0.0194 \\
  ULMFiT & 1k-1k   & 0.4193 & 0.8183 & 0.8236 & +0.0835 & +0.0782 \\ 
  ULMFiT & 100-100 & 0.7020 & 0.4390 & 0.4904 & +0.4628 & +0.4114 \\
 \hline
\end{tabular}
\caption{\label{finetuning} Finetuning Results. 10K Val Acc refers to validation accuracy when evaluating on the validation set of the original 10K-10K split. Err Increase refers to the increase in error when number of training examples were reduced to a particular split. 10K Err Increase refers to the increase in error when evaluating on the original 10K-10K split validation set once training examples are reduced. * pertains to instances when the 10K Val Acc is higher than the Val Acc of a particular split.}
\end{table*}

\subsection{Discussion}

We can see that language model pretraining can aid in low-resource settings as empirically shown in the experiments above. The finetuned models were shown perform consistently even when the number of training examples were reduced by evaluating on the same validation set. 

ULMFiT performed marginally better than BERT (a difference of 0.0201) when finetuned on the full dataset. ULMFiT has the advantage that it requires less computational power and resources to effectively train end-to-end. An AWD-LSTM language model can be trained in a relatively-modern GPU and can be finetuned with relative speed to BERT. This makes ULMFiT ideal in most low-resource cases when pretrained models are unavailable as it is cheaper to produce AWD-LSTM language models than pretrained BERT models. 

On the other hand, it is worth to note that BERT performed more consistently on average than ULMFiT. BERT experienced a lower error increase on average compared to ULMFiT, with BERT-Cased, BERT-Uncased, and ULMFiT experiencing an average validation error increase of 0.0719, 0.1201, and 0.1840, respectively. BERT is also more resilient to drastic reduction in training examples. When reducing the splits from 1K-1K to 100-100, BERT (evaluated on the full 10K-10K split validation set) experienced an increase of error by 0.192 on the cased models and 0.2043 on the uncased models. ULMFiT, on the other hand, experienced an error increase of 0.3332.

BERT also has the advantage of being bidirectional, which allows it to look at both left and right context as needed, compared to ULMFiT, where the AWD-LSTM language model only used left context. BERT is also significantly more deep than ULMFiT's AWD-LSTM, with BERT-Base having 12 layers and 12 attention heads as opposed to an AWD-LSTM's 3 layers. This allows it to learn more complex relationships within the data.

While the advantages of the much-larger BERT are evident, it is important to note that it requires compute resources orders of magnitude greater than needed when training an AWD-LSTM. Pretraining BERT requires at least a TPU in order to meet the memory requirements. It also takes much longer to train than an AWD-LSTM. The pretrained models in this work are all BERT-Base models, using one whole TPU in order to train and at least a little over a day to achieve robust results. Larger datasets and model configurations would naturally require more time and memory. This makes scaling up to BERT-Large hard. The original BERT implementation used 4 cloud TPUs for BERT-Base and 16 TPUs for BERT-Large. Finetuning BERT likewise has sizeable memory requirements, with BERT-Base requiring at least a modest-to-high-end GPU. BERT-Large will have a difficulty in GPU-finetuning\footnote{https://github.com/google-research/bert\#out-of-memory-issues}, requiring the use of gradient accumulation and other techniques to simulate larger batch sizes as small batch sizes will hurt finetuning performance. While powerful, the resources needed to use BERT make it restrictive.

\section{Conclusion}
We show that language model finetuning methods aid in low-resource settings, especially when the number of expert-annotated examples is scarce. 

Language model pretraining offers two advantages: first, performing pretraining only requires unlabeled text corpora, which is virtually abundant even in low-resource settings. Second, once pretraining is done, finetuning is inexpensive and can be performed multiple times on the same pretrained model. This allows researchers to use only a fraction of resources to create robust baselines even in low-resource settings.

Choosing the finetuning technique involves a cost-consistency tradeoff. We propose the use of ULMFiT as a general-case finetuning-based baseline as it's pretraining step is relatively less expensive than BERT. While BERT is powerful, it's compute and memory requirements make it restrictive, and should only be used if a pretrained model exists or if the resources available permit it's use.

\section*{Acknowledgments}
The authors would like to thank the TensorFlow Research Cloud (TFRC) program, which allowed the pretraining of BERT models more accessible. \\

\bibliography{emnlp-ijcnlp-2019}
\bibliographystyle{acl_natbib}

\end{document}